\documentclass[journal]{IEEEtran}

\usepackage{amsmath,bm}
\usepackage{graphicx}
\usepackage{amsmath,amssymb} %
\usepackage{color}
\usepackage{subfigure}
\usepackage{algorithm}
\usepackage{algpseudocode}

\usepackage{color}
\usepackage{epsfig}
\usepackage{graphicx}
\usepackage{caption}
\usepackage{amsmath}
\usepackage{amssymb}
\usepackage{enumerate}

\usepackage{booktabs}
\usepackage{adjustbox}
\usepackage{multirow}

\usepackage{amsfonts}
\usepackage{multirow}
\usepackage{subfigure}
\usepackage{color}
\graphicspath{{./figures/}}
\usepackage{epstopdf}
\usepackage{float}
\usepackage{algorithmicx}

\usepackage{multirow}

\usepackage{algorithm}

\ifCLASSINFOpdf
\else
\fi

\begin{document}
\title{Aggregation Signature for Small Object Tracking}

\author{Chunlei~Liu,
        Wenrui~Ding,
        Jinyu~Yang,
        Vittorio Murino, ~\IEEEmembership{Senior Member, IEEE,}
        Baochang Zhang,~\IEEEmembership{Member~IEEE,}
        Jungong~Han,~\IEEEmembership{Member~IEEE,}
        Guodong~Guo,~\IEEEmembership{Senior Member, IEEE,}%
\thanks{C. Liu is with School of Electrical and Information Engineering, Beihang University, Beijing, China.
E-mail: liuchunlei@buaa.edu.cn}
\thanks{W. Ding is with Unmanned System Research Institute, Beihang University, Beijing, China.
E-mail: ding@buaa.edu.cn}
\thanks{J. Yang is with School of Computer Science in the University of Birmingham, British.
E-mail:  yangjinyu@buaa.edu.cn}
\thanks{V. Murino is with University of Verona, Verona, Italy, and is also with Pattern Analysis and Computer Vision department at the Istituto Italiano di Tecnologia, Genoa, Italy.
E-mail: vittorio.murino@iit.it}
\thanks{B. Zhang is with School of Automation Science and Electrical Engineering, Beihang University, Beijing, China.
Baochang Zhang is also with Shenzhen Academy of Aerospace Technology, Shenzhen, China. E-mail: bczhang@buaa.edu.cn}
\thanks{J. Han is with the WMG Data Science Group, University of Warwick,
Coventry CV4 7AL, U.K. (e-mail: jungonghan77@gmail.com)}
\thanks{G. Guo is with Institute of Deep Learning, Baidu Research and National Engineering Laboratory for Deep Learning Technology and Application. E-mail: guoguodong01@baidu.com}
\thanks{Wenrui Ding and Baochang Zhang are the corresponding authors.}
}

\markboth{TRANSACTIONS ON IMAGE PROCESSING}%
{Shell \MakeLowercase{\textit{et al.}}: Bare Demo of IEEEtran.cls for IEEE Journals}

\maketitle

\begin{abstract}
Small object tracking becomes an increasingly important task, which however has been largely unexplored in computer vision. The great challenges stem from the facts that: 1) small objects show extreme vague and variable appearances, and 2) they tend to be lost easier as compared to normal-sized ones due to the shaking of lens. In this paper, we propose a novel aggregation signature suitable for small object tracking, {especially aiming for the challenge of sudden and large drift}. We make three-fold contributions in this work. First, technically, we propose a new descriptor, named aggregation signature, based on saliency, able to represent highly distinctive features for small objects. Second, theoretically, we prove that the proposed signature matches the foreground object more accurately with a high probability. Third, experimentally, the aggregation signature achieves a high performance on multiple datasets, outperforming the state-of-the-art methods by large margins. Moreover, we contribute with two newly collected benchmark datasets, i.e., small90 and small112, for visually small object tracking. The datasets will be available in https://github.com/bczhangbczhang/.
\end{abstract}


\IEEEpeerreviewmaketitle

\section{Introduction}
\IEEEPARstart{W}{hile}  several tracking methods have been developed over the past decade \cite{Stalder2010Cascaded,Heber2013Segmentation,Kalal2012Tracking,Danelljan2017Discriminative,Hare2016Struck,Zhang2016Bounding,T-CYB} and have been proven to be successful in many applications, such as robotics or video surveillance, %
tracking small objects in videos  still remains a challenging problem, in particular when the complex scenarios and real time constraints are to be considered.
In this paper, small objects mean that the targets in images have sizes of less than 1\% of the whole image. The challenge of small object tracking mainly roots in two main facts: first, the visual features of small objects are extremely fickle, thus making feature representation difficult; {second, sudden and large drift always occurs to small objects in tracking because of the shaking of the lens, compared to the normal-sized objects. The so-called sudden and large drift is that the target distance between two adjacent frames in the image coordinate system is two times larger than the target size.}

For a long time, researchers only reported tracking results on common benchmarks using reasonably sized targets, but paid less attention to the  small-object tracking problem. Just few existing algorithms related to the small object tracking, while that were designed to enhance the visual features of such a type of targets, with the hope that tracked objects would no longer be lost if robust features were exploited. For instance, the method in \cite{Ahmadi2016Small,SPL}  integrates both spatial and frequency domain features in order to localize the targets more accurately. Alternatively, the method in \cite{Ahmadi2015Small} tends to enhance the robustness of a tracker by strengthening the feature representations (e.g., target attributes) for the small targets. Recently, Rozumnyi et al. \cite{Rozumnyi2017The} have proposed to deal with fast moving and motion blur problems of the objects, but the performance is unsatisfactory due to low resolution and complex background clutters. Regarding now deep learning methods \cite{liu2019circulant}\cite{liu2019RBCN} are developed, we think the high-level features seem not to be effective for small objects. Moreover, we doubt that a continuous tracking of small-sized objects can be guaranteed even if robust visual features are exploited, considering the fact that small targets can easily be confused with the noise and clutters in real scenes.
In other words, it might be more
realistic to allow small objects to get lost during tracking,
while investigating a better solution to re-detect them.

The intuition  here is about ``how human beings recognize the small target when it is lost due to clutter background?''  Most likely, humans first look at the salient objects/regions popping up in the scene, and further verify whether one of the salient objects is the target of interest \cite{Choi2017Attentional}. A few works mimic human being's behavior and involve the saliency information in object tracking.
For example, the method in \cite{Ma2017A} integrates saliency for the representation of context, while \cite{Yi2014Initialization,Fan2010Discriminative,Luo2013Salient} incorporate saliency into appearance models in various ways in order to improve the robustness of the tracker.  However, as they mostly focus on the target appearances in the image domain, performance is not satisfactory since the appearance is implicitly weak for small objects.  Therefore, they might only be reliably applied for tracking normal-sized objects.
In this paper, we propose a new saliency online learning framework, termed aggregation signature, and focus on small object tracking.
To the best of our knowledge, no saliency-based methods have utilized all context information, including intensity, saturation, saliency and motion information, for small object tracking yet.

\begin{figure*}[!t]
\centering
\includegraphics[scale=0.4]{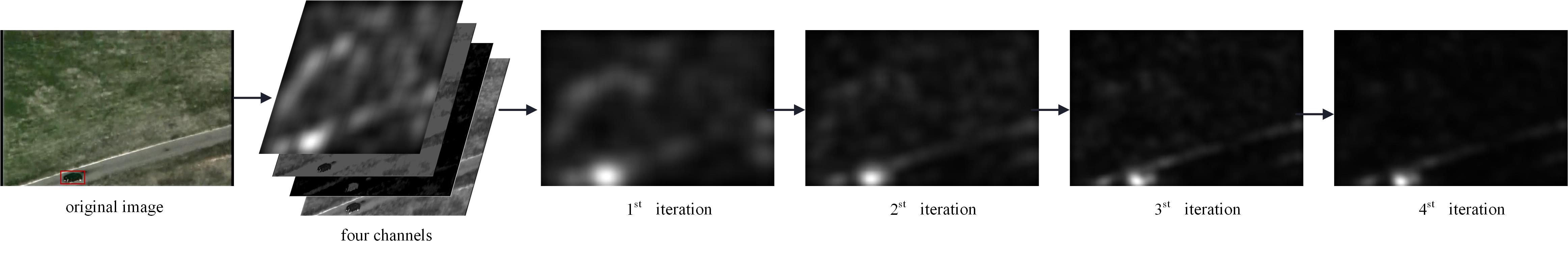}
\caption{The aggregation signature results are shown at different iterations, reflecting that the tracked target becomes more salient in the learning process. Our aggregation signature constitutes the first attempt to incorporate the tracked target information into the quaternion discrete cosine transform (QDCT) image signature,  whose  aggregation capacity is proved in theoretical terms. }
\label{iteration}
\end{figure*}

Unlike handcrafted image signatures, which are simple yet powerful tools to spatially match the sparse foreground objects in an image \cite{Hou2012Image, Schauerte2012Quaternion}, the explicit advantage  of our aggregation signature lies in a learning mechanism exploited to build %
an adaptive target signature.
The result is that it can quickly detect the salient objects even though they are very small, which can further improve the (re-)localization performance of the trackers.
  We open up  a new direction to track small objects by mimicking the human attention mechanism. %
 In particular, the theoretical evidence proves that it is more effective, and that the resulting foreground saliency map from our aggregation signature becomes more consistent with the target appearance along iterations, as shown in Fig. \ref{iteration}.  Moreover, the aggregation signature is so generic that it can be integrated into other trackers. In summary, the contributions of this paper include:

(i)  The proposed aggregation signature is proved, in the theoretical terms, to be more efficient for sparse foreground detection, makings the tracked target more salient as compared to the background.

(ii) The aggregation signature improves the capacity of accumulating information for the target  based on a learning mechanism, whereas the conventional image signatures are handcrafted and more likely prone to fail to adapt to the target.

(iii) New challenging datasets -- small90 and small112 -- are  collected for small object tracking evaluation. The datasets are publicly available for further research development.

\section{Aggregation Signature}\label{sec:Aggregation Signature Properties}
Image signature is a simple yet powerful tool to spatially match the sparse foreground of an image \cite{Hou2012Image}. By using the sign function of DCT, the resulting handcrafted descriptor can approximately detect salient image regions efficiently. Rather than separating a color image into three channel images and computing image signatures respectively, QDCT \cite{Schauerte2012Quaternion} can discriminate the relative importance of  four components by introducing a quaternion component. In general, both DCT and QDCT based image signatures are handcrafted methods with no involvement of a learning process.
Differently, the proposed aggregation signature improves the discriminative capability of QDCT signature via \emph{learning} multi-cue information, in particular the target prior information.

\subsection{Definition of Aggregation Signature}
We begin by considering an image ${\rm x}$ which exhibits the following structure:
\begin{align}
  {\rm \mathbf x=\mathbf f+\mathbf b,
  \rm\quad \mathbf x,\mathbf f,\mathbf b}\in \rm{{{\rm \mathbb R}}^\emph{N}},
\end{align}
where ${\rm \mathbf f}$ represents the foreground and ${\rm \mathbf b}$  represents the background.
Please refer to Table \ref{1} for the definitions used throughout the rest of this section.
Formally, the aggregation signature (AS) is defined as:
\begin{align}
 \rm AS\left(\mathbf x^ \emph{i}_ \emph{Q}\right)=sign\left(QDCT\left(\mathbf x^\emph{i}_ \emph{Q} \right)\right),
\end{align}
where $\rm sign(\cdot)$ is the entrywise sign operator, $i$ represents the ${ i^{th}}$ iteration and $ Q$ represents the 4 channels in use.
Then, the reconstructed image can be defined as:
\begin{align}
 \rm \bar{\mathbf x}^\emph{i}_ \emph{Q}=IQDCT\left(AS\left(\mathbf x^\emph{i}_ \emph{Q} \right)\right),
\end{align}
and
\begin{align}
\rm \mathbf{x}^\emph{i}_ \emph{Q}= \bm {\pi}\circ \mathbf{S}^\emph{i}+\mathbf{I_1}{\emph{j}}+\mathbf{I_2}{\emph{k}}+\mathbf{I_3}{\emph{h}},
\end{align}
where $\rm \mathbf S^\emph{i}=\bar{\mathbf x}^{\emph i-1}_ \emph{Q} \circ \widetilde{\bar{\mathbf x}}^{\emph i-1}_ \emph{Q}$ \footnote{If $\rm \mathbf S^\emph{i}$ is the image signature based on DCT, we have $\rm \mathbf S^{\emph i}=\bar{\mathbf x}^{\emph i-1}$.}, ${\rm \bar{\mathbf x}^\emph i_Q}$ represents the reconstructed result in the ${\emph i^{th}}$ iteration with ${\rm \widetilde{\bar{\mathbf x}}^\emph{i}_\emph{Q}}$ as its conjugate form, and $\rm \circ$ represents the element wise product. ${\rm \mathbf {I_1}}$, ${\rm \mathbf {I_2}}$, ${\rm \mathbf {I_3}}$ represent
three different channels such as any one channel of RGB, image intensity and image saturation (or motion  in tracking). $\rm \bm \pi$ is a two-dimensional prior related to the tracked target, which will be elaborated in  Section \uppercase\expandafter{\romannumeral4}.
\subsection{Foreground Aggregation Signature Properties}
In this section, we provide evidence that, for an image which adheres to a certain mathematical structure, the background can be suppressed by the aggregation signature.

\textbf{Proposition:} The image reconstructed from the aggregation signature matches the foreground object more accurately in the  learning process with a high probability as follows:
\begin{align}
\begin{split}
&\rm \mathcal P\left( \emph E \left( \frac{<\mathbf f,\mathbf S^{\emph i+1}>}{\|\mathbf f\| \cdot \|\mathbf S^{\emph i+1}\|} \right)-
      \emph E \left( \frac{<\mathbf f,\mathbf S^\emph {i}>}{\|\mathbf f\| \cdot \|\mathbf S^\emph {i}\|} \right)> 0\right)\\
      &> \rm \left(1-2\varepsilon\left(1-\varepsilon\right)\right)^\emph N,
\quad for
\quad \forall \emph i\ge1,
\end{split}
\end{align}
where $\rm \mathcal P$ stands for probability, $\rm \varepsilon$ is a  small positive value, $\rm \emph N$ represents total image pixel number, $\rm \|\cdot\|$ represents the $l^2$ norm, $\rm <\cdot,\cdot>$ denotes the inner-product.  ${\rm \emph E}\left(\cdot\right)$ denotes expectation, which reveals about the similarity between the foreground and the object saliency information obtained by aggregation signature.

\begin{table}[!t]
\begin{center}
\caption{Notation and terms used in this paper.}
\label{table:headings}
\begin{tabular}{ll}
\hline\noalign{\smallskip}
Terms & Notation\\
\noalign{\smallskip}
\hline
\noalign{\smallskip}
$\widehat{\mathbf x}$&DCT($\mathbf x$).\\
$\rm sign(\mathbf x)$& The entrywise sign operator.\\
$\rm {\widetilde{\mathbf x}}$& The conjugate form of $\rm \mathbf x$.\\
$\bar{\mathbf x}$ & $\rm IDCT\left[sign\left(\widehat{\mathbf x}\right)\right]$, the reconstructed image of DCT.\\
$\rm \bar{\mathbf x}_Q$ & $\rm IQDCT\left[sign\left(QDCT\left(\mathbf x\right)\right)\right]$,the reconstructed image\\
$$& of QDCT.\\
${E}\left(X\right)$ &The expectation of random variable $X$.\\
$\rm \|\mathbf x\|_\emph p$ &The $l^p$ norm of vector $\rm \mathbf x$. ($\emph p$=2 if omitted).\\
$\rm <\mathbf x,\mathbf y>$&The inner-product of $\mathbf x$ and $\mathbf y$.\\
$\rm \circ$&The Hadamard (entrywise) product operator. \\
$\rm \Omega_\mathbf x$& Support set of $\widehat{\mathbf x}$.\\
\hline
\end{tabular}
\label{1}
\end{center}
\end{table}

\textbf{Proof:} We know the transform between QDCT and DCT is
\begin{align}
\begin{split}
  \rm QDCT\left(\mathbf x^{\emph i+1}_Q\right)
\rm &=\sqrt{1/3}\rm DCT\left(-\mathbf{I_1}-\mathbf{I_2}-\mathbf{I_3}\right)\\
&+ \sqrt{1/3}\rm DCT\left(\mathbf{I_3}+\bm \pi\circ \mathbf S^\emph i -\mathbf{I_2}\right) \emph{j}\\
& \rm + \sqrt{1/3}DCT\left(\mathbf{I_1}+\bm \pi\circ \mathbf S^\emph i -\mathbf{I_3}\right) \emph{k}\\
&\rm + \sqrt{1/3}DCT\left(\mathbf{I_2}+\bm \pi\circ \mathbf S^\emph i -\mathbf{I_1}\right) \emph{h}.
\end{split}
\end{align}

For ease of explanation, we only focus on one channel, that is to say $\rm \mathbf S^{\emph i}=\bar{\mathbf x}^{\emph i-1}$ and the result can be easily generalized for the quaternion case in a straightforward way, then we have
\begin{align}
\begin{split}
\rm  \|\overline{\mathbf x}^{\emph i+1}\|^2&=\rm IDCT<\rm\left(sign\left(\widehat{\mathbf x}^{\emph i+1}\right)\right), \left(sign\left(\widehat{\mathbf x}^{\emph i+1}\right)\right)>\\
&=\rm \sum\limits_{\emph p }\left(sign\left(\widehat{\emph x}^{\emph i+1}\left(\emph p\right) \right) \right)^2,\\
\end{split}
\end{align}
where $\rm \widehat{\mathbf x}=DCT(\mathbf x)$ and $p$ represents the points of the corresponding support set. We note that the proof is applicable to channels $j,k,h$ in Equ. (6), so we take the channel $j$ for example.
Then, we have
\begin{align}
\begin{split}
\rm  sign\left(\widehat{\mathbf x}^{\emph i+1}\right)&=\rm sign\left(DCT\left(\mathbf{I_3}+\bm \pi \circ \bar{\mathbf x}^\emph i-\mathbf{I_2}\right)\right)\\
&\rm =sign\left(DCT\left(\mathbf{I_3} -\mathbf{I_2}\right)+\bm \pi\circ sign\left(\widehat{\mathbf x}^\emph i\right) \right).
\end{split}
\end{align}

Since the results obtained by DCT are independent of each other, we assume
\begin{align}
\begin{split}
{\rm{\mathcal P}}\left(\widehat  x\left(\emph p\right)=\pi\left(\emph p\right) \right)&= {\rm{\mathcal P}}\left(\widehat x\left(\emph p\right)=0\right)\\
&= {\rm{\mathcal P}}\left(\widehat x\left(\emph p\right)=-\pi\left(\emph p\right) \right)=\rm \varepsilon,
\label{varep}
\end{split}
\end{align}
where $\varepsilon$ is very small, since the probability that the DCT output is equal to a certain value is very small. Then we have the following statement:
\begin{align}
\begin{split}
\rm \mathcal P\left(\|\overline{\mathbf x}^\emph i\|^2=\|\overline{\mathbf x}^{\emph i+1}\|^2\right)> \rm \left(1-2\varepsilon\left(1-\varepsilon\right)\right)^\emph N,
\end{split}
\end{align}
which means that  in a high probability we have $\rm \|\overline{\mathbf x}^\emph i\|=\|\overline{\mathbf x}^{\emph i+1}\|$, considering that $\rm \varepsilon$ is very small.

Similarly, we have
\begin{align}
\begin{split}
&\rm <\overline{\mathbf f},\overline{\mathbf x}^{\emph i+1}>=
\rm < sign\left(\widehat{\mathbf f}\right), sign\left(\widehat{\mathbf x}^{\emph i+1}\right)>\\
&\rm = < sign\left(\widehat{\mathbf f}\right), sign\left(DCT\left(\mathbf{I_3} -\mathbf{I_2}\right)+\bm \pi\circ sign\left(\widehat{\mathbf x}^\emph i\right) \right)>.
\end{split}
\end{align}

Since $\pi \ge 0$, if $\rm \vert DCT\left({\emph I_3}\left(\emph p\right)-{\emph I_2}\left(\emph p\right)\right)\vert >  \pi\left(\emph p\right) $,
then we have
\begin{align}
\begin{split}
&\rm sign\left(DCT\left({\emph I_3}\left(\emph p\right) - {\emph I_2}\left(\emph p\right)\right) \right)-\pi\left(\emph p\right) sign\left(\widehat{\emph  x}^\emph i\left(\emph p\right)\right)\\
 &\rm < sign\left(DCT\left({\emph I_3}\left(\emph p\right) -{\emph I_2}\left(\emph p\right)\right)+ \pi\left(\emph p\right) sign\left(\widehat{\emph  x}^\emph i\left(\emph p\right)\right) \right) \\
&\rm < sign\left(DCT\left({\emph I_3}\left(\emph p\right) -{\emph I_2}\left(\emph p\right)\right) \right)+\pi\left(\emph p\right) sign\left(\widehat{\emph  x}^\emph i\left(\emph p\right)\right).
\end{split}
\end{align}

Combining (11) and (12), we have
\begin{align}
\begin{split}
&\rm <\overline{\mathbf f},\overline{\mathbf x}^{\emph i+1}>
\rm = < sign\left(\widehat{\mathbf f}\right), sign\left(\widehat{\mathbf x}^\emph i+DCT\left(\mathbf{I_3} -\mathbf{I_2}\right)\right)> \\
&\rm \geq <  sign\left(\widehat{\mathbf f}\right),\bm \pi\circ sign\left(\widehat{\mathbf x}^\emph i\right)-sign\left(DCT\left(\mathbf{I_3} -\mathbf{I_2}\right)\right)>\\
&\rm = <\overline{\mathbf f},\overline{\mathbf x}^{\emph i}>+< sign\left(\widehat{\mathbf f}\right), sign\left(DCT\left(\mathbf{I_2} -\mathbf{I_3}\right)\right)>.
\end{split}
\end{align}

Based on the image signature proposed by Hou \cite{Hou2012Image}, we have
\begin{align}
\begin{split}
&\rm < sign\left(\widehat{\mathbf f}\right), sign\left(DCT\left(\mathbf{I_2} -\mathbf{I_3}\right)\right)>
\\&\geq
\rm  |\Omega_\mathbf f|-|\Omega_{{\mathbf{I_2}}-{\mathbf{I_3}}-\mathbf f}|,
\end{split}
\end{align}
where $\rm \Omega_\mathbf x$ represents the support set of $\rm \widehat{\mathbf x}$.
Given the bound  $\rm |\Omega_\mathbf f|\geq \frac{2}{3}\emph N$ \cite{Hou2012Image}, we have
\begin{align}
\begin{split}
\rm E\left(<sign\left(\widehat{\mathbf f}\right),sign\left(DCT\left(\mathbf{I_2}-\mathbf{I_3}\right) \right)> \right) \geq \frac{1}{3}\emph N.
\end{split}
\end{align}

And then it becomes
\begin{align}
\begin{split}
\rm E\left(<\overline{\mathbf f},\overline{\mathbf x}^{\emph i+1}>-<\overline{\mathbf f},\overline{\mathbf x}^{\emph i}> \right) \geq \frac{1}{3}\emph N.
\end{split}
\end{align}

For a spatially sparse foreground, we have the following statement:
\begin{align}
\rm E\left(<\mathbf f,\overline{\mathbf x}^{\emph i+1}>-<\mathbf f,\overline{\mathbf x}^{\emph i}> \right) > 0.
\end{align}

Together with Equ. (10), we have
\begin{align}
\begin{split}
 & \rm  \mathcal P\left( E \left( \frac{<\mathbf f,\overline{\mathbf x}^{\emph i+1}>}{\|\mathbf f\| \cdot \|\overline{\mathbf x}^{\emph i+1}\|} \right)-
      E \left( \frac{<\mathbf f,\overline{\mathbf x}^{\emph i}>}{\|\mathbf f\| \cdot \|\overline{\mathbf x}^{\emph i}\|} \right)> 0\right)\\
      &> \left(1-2\varepsilon\left(1-\varepsilon\right)\right)^\emph N,
\end{split}
\end{align}
which proves the proposition.

\textbf{Remark:} Here, $\varepsilon$ is very small as in Equ. (\ref{varep}), e.g., $\varepsilon = 0.0001$, and the probability mentioned above is $81$\% when $N = 1024$. In other words,  background is suppressed more during learning aggregation signature with high probability. We also did a statistic analysis on $\varepsilon$ in Equ. (\ref{varep}) based on the MSRA-B dataset \cite{Liu2007Learning} , which indicates that $\varepsilon$ is very small less than $\varepsilon= 1.5e-9$.

\section{Aggregation Signature Tracker}\label{sec:Aggregation Signature Tracker}

\begin{figure*}[!t]
\centering
\includegraphics[scale=0.5]{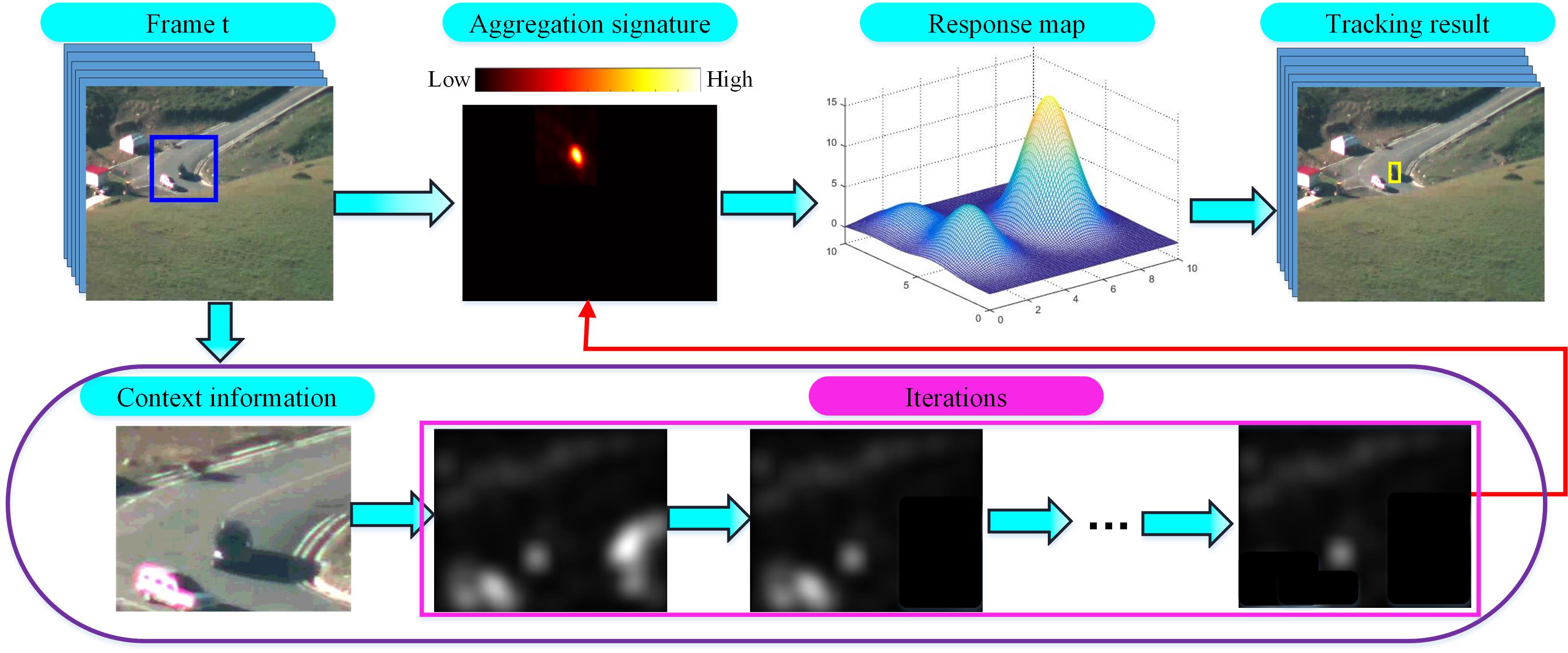}
\caption{Scheme of the aggregation signature tracker, which includes the base tracker and re-detection stages, particularly for small objects. The part of aggregation signature calculation illustrates the saliency map calculation in the re-detection procedure. Once a drifting is detected, we choose the search region around the center of the previous target location to calculate the saliency map via aggregation signature. The blue box is the search region. In the learning process, the target prior ($\bm \pi$) and the context information in the blue box are used to learn the saliency map that helps to find a new initial position, where the base tracker will be performed again for re-detection.}
\label{fig:1}
\end{figure*}

We exploit the aggregation signature to enhance the re-detection process for small object tacking, which is called aggregation signature tracker (AST).
More specifically, when a target is found drifting by a thresholding method, a saliency detection with the tracked target as prior will be triggered, which enables the online aggregation signature to suppress the background data. Together with the context information indicated in different channels, we re-detect the objects to relocate the tracked target. The whole tracking procedure is illustrated in Fig. \ref{fig:1}(a) and Algorithm 1, and we elaborate each key component in the following.

\textbf{Drifting detection:} As evident on output constraint transfer tracking method (OCT) \cite{Zhang2017Output},
 a simple distribution is necessary and significant to achieve high efficiency. OCT builds upon a reasonable assumption that the response to the target image follows a Gaussian distribution, so we trigger the re-detection process based on a thresholding method as:
\begin{align}
\ |r - \mu|>T_g,
\end{align}
where $ \mu$ represents the mean response using all previous frames, $r$ represents the maximum response of the current frame, and $T_g$ is the threshold. The target is supposed to be lost if the response of the current frame is far from the average response. Once the target is occluded or out of view, this mechanism helps us search continuously in the following frames.

\textbf{Saliency map calculation:} The aggregation signature is used to obtain the saliency map and to further coarsely re-localize the target. Through $ R$ iterations, we gradually smooth the aggregation signature by a Gaussian kernel \cite{Hou2012Image} to obtain the saliency map.
The salient regions are regarded as the coarse candidate positions of the target, on which a re-detection process is performed still based on the selected base trackers.
It should be mentioned that involving the targeted object, as a prior in saliency detection, does not occur in the conventional methods. Two key components are elaborated as follows:
\begin{algorithm} [!t]
\caption{\textbf{- Aggregation signature tracker}}
	\label{alg:proposed tracking algorithm}
	\begin{algorithmic}[1]
		\State  Initial target bounding box ${\bf{b}}^1=[p_x,p_y,w,h] $
        \State  Initial $ \xi=1$, $ R=4$, $ M=6$
        \If{the frame $ t<=3$}
		\Repeat
		\State Crop out the search windows  according to $ \bf{b}^{\emph t}$, and extract feature
		\State Compute the maximal response according to base tracker
		\State The position is obtained according to the maximal response $ r$
		\State Updating essential parameters of the base tracker
        \State $t=t+1$
		\Until{ $t$}==3
		\EndIf
		\State Compute the mean $\mu$ of response using all previous frames
        \If{the frame $t>3$}
		\Repeat
		\State Crop out the search windows  according to $ \bf{b}^{\emph t}$, and extract feature
		\State Compute the maximal response according to base tracker
        \If{$|r-\mu|>T_g$ }
		\State Crop out the target search regions
        \State Obtain channels $\rm \mathbf I_1, \mathbf I_2,\mathbf  I_3 ,\mathbf S$ according to channels design in section 4
		\State Calculate the aggregation signature saliency map as illustrated in section 3.1
   		\State Obtain the coarse target location based on the saliency map
  		\State Compute new target location according to base tracker
  		\EndIf
        \State $t=t+1$
		\Until{ the end of the video}		
        \State Updating parameters of the base tracker
    	\State Updating $\mu$
       	\EndIf
		\State \textbf{end}
	\end{algorithmic}
\end{algorithm}

\textbf{1) Channels design:} We denote the input image captured at frame $\emph t$ as $\rm  \mathbf F^\emph t$, where $\rm \mathbf R^\emph t$, $\rm \mathbf G^\emph t$, and $\rm \mathbf B^\emph t$ are the red, green and blue channels of $\rm \mathbf F^\emph t$.
Then, we obtain three channels used in our aggregation signature representing as: intensity $ \mathbf  I_1^\emph t=(\mathbf R^\emph t+\mathbf G^\emph t+\mathbf B^\emph t)/3$, saturation $\rm \mathbf I_2^\emph t=max(\mathbf R^\emph t,\mathbf G^\emph t,\mathbf B^\emph t)$ and movement $\rm \mathbf I_3^\emph t=|\mathbf I_1^\emph t-\mathbf I_1^{\emph t-\tau}|/3$, respectively, where $\tau$ is a constant.  We deploy  image signature \cite{Hou2012Image} to calculate the initial saliency map as the first channel $\rm \bm\pi\circ \mathbf S^{\emph i}$.

\textbf{2) Target Prior:}
As shown in Fig. \ref{fig:1} (b), we select $M$ salient regions similar to the target in the last frame in size. Next,  we assign each candidate a weight indicating the similarity to a target prior information, which is measured simply by the Euclidean distance as:
\begin{align}
{\pi} _{\emph n}^{\emph t}=\frac{1}{\sqrt{2\pi }\xi }{{e}^{-\frac{1-{{d}_{\emph n}^\emph t}}{2{{\xi }^{2}}}}},
\end{align}
where $ \pi _{\emph n}^{\emph t}$ denotes   the weight of the ${{n}^{th}}$  region for the candidate saliency map at the $t^{th}$ frame, $ \xi$ is a constant.  $ d_{\emph n}^{\emph t}=\sum\limits_{\text{\emph i}=\text{1}}^{\text{255}}{\sqrt{{{(\mathbf H^{t}\left(\emph i\right)-\mathbf y_{\emph n}^{\emph t}\left(\emph i\right))}^{2}}}}$,
where $\mathbf y_\emph n^\emph t$ represents the histogram of the candidate saliency map, while $\rm\mathbf H^\emph t$ denotes the target  histogram for the $t^{th}$ frame  calculated by
\begin{align}
 \rm\mathbf H^\emph t = \sigma \rm\mathbf H^\emph t + (1-\sigma) \rm\mathbf H^\emph 1 ,
\end{align}
where $\sigma$ is 0.5 in this paper.
We note that the weights are set to $1$ for the regions outside the selected salient areas.

\section{Experiments}
In this section, we  evaluate the aggregation signature based on our small90 dataset and a visual saliency benchmark MSRA-B \cite{Liu2007Learning}. We further test the performance of our aggregation signature based tracker on the small90, small112, UAV123\_10fps \cite{Mueller2016A} and UAV20L \cite{Mueller2016A} according to the object tracking benchmark \cite{Wu2015Object}.
The test platforms are Intel I7 2.7 GZ (4 cores) CPU with 8G RAM, and GPU with NVIDIA GeForce GTX 1070.

\begin{figure}[!t]
 \centering
\includegraphics[scale=0.65]{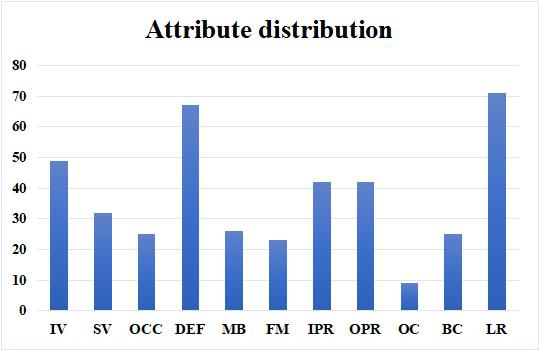}
\caption{Attribute distribution across small90.}
\label{canshu}
\end{figure}

\begin{figure*}[!t]
 \centering
\includegraphics[scale=1.0]{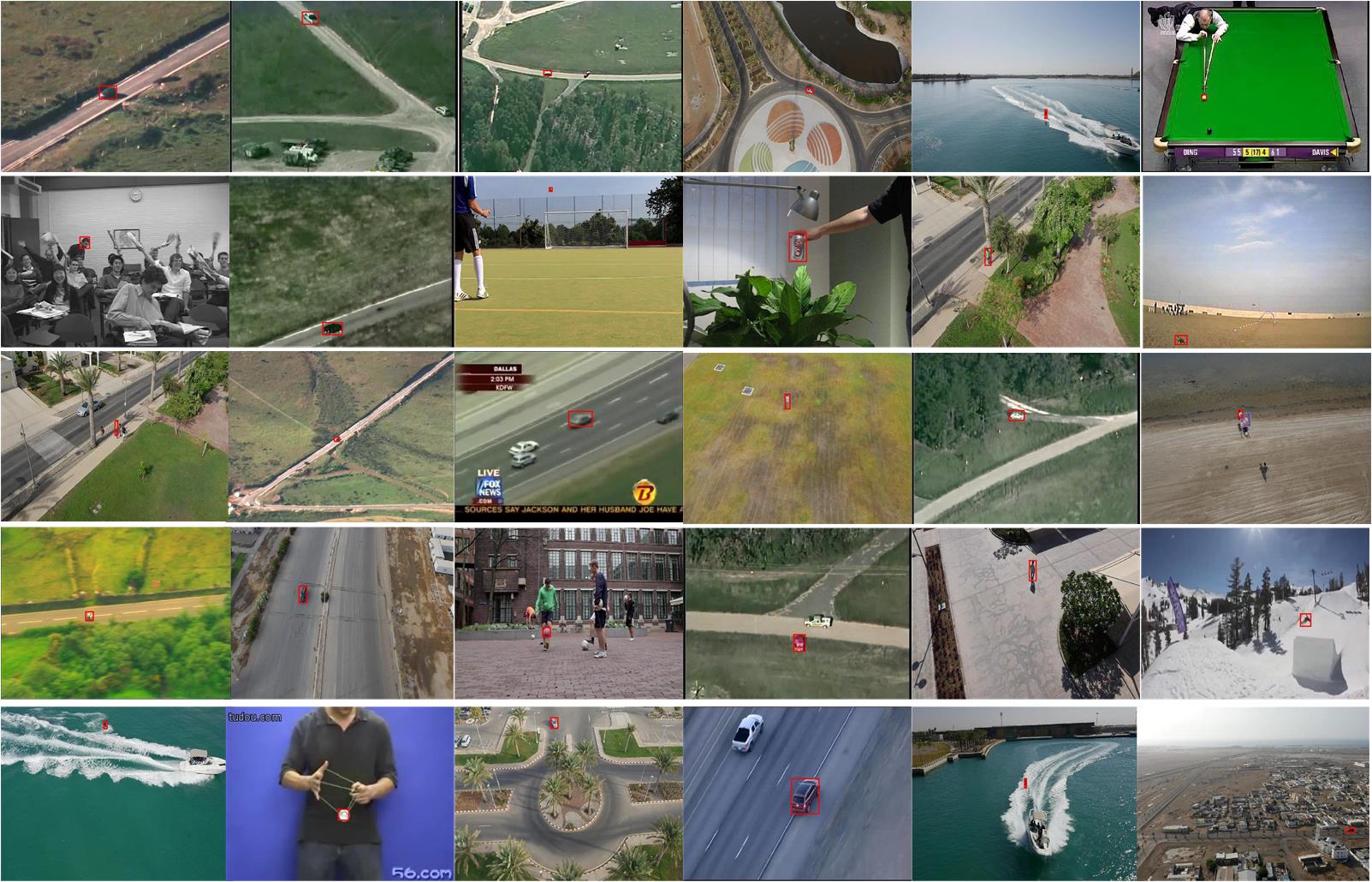}
\caption{The first frames of selected sequences from small90. The red bounding box indicates the ground truth.}
\label{firstframe}
\end{figure*}
\vspace{-5 mm}

\subsection{Datasets}
Few datasets are available for small object tracking task. We establish a comprehensive database, termed small90 benchmark, consisting of 90 annotated small-sized object sequences, where several additional challenges, such as target drifting and low resolution, have been encompassed.  We add 22 more challenging sequences into small90, and obtain another new dataset termed as small112. Each sequence is categorized with 11 attributes - illumination variations (IV), scale variations (SV), occlusions (OCC), deformations (DEF), motion blur (MB), fast motion (FM), in-plane rotation (IPR), out-of-plane rotation (OPR), out-of-view (OV), background clutters (BC) and low resolution (LR), for better analysis of the tracking approaches.  The
 attribute distribution in our dataset is plotted in Fig. \ref{canshu}, which shows that some attributes occur more frequently, e.g., LR, than the others. We note that one sequence is often annotated with multiple attributes. The examples of first frames from our datasets are illustrated in Fig. \ref{firstframe}.

\subsection{Aggregation Signature on Image}
\begin{figure*}[!t]
\centering
\includegraphics[scale=0.37]{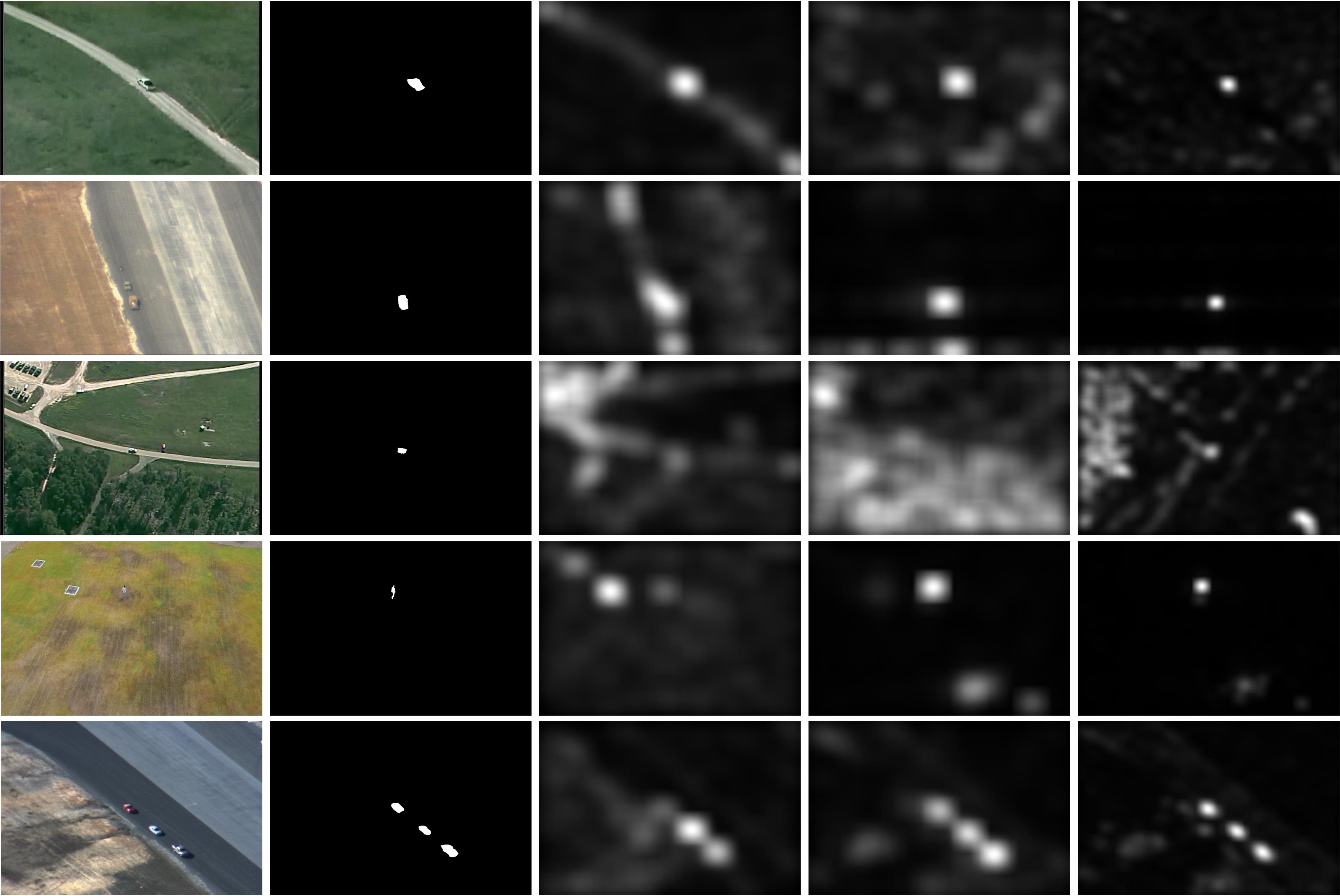}
\caption{Representative results of different signature methods. For each group from left to right, they are the original image, ground truth, IS, QIS and AS, respectively. Comparing with the other methods, AS yields the best background suppression performance.}
\label{xianzhutu}
\end{figure*}

We first evaluate how aggregation signature can enhance the performance of saliency detection, based on the commonly used metrics including location-based metrics normalized scanpath saliency (NSS) \cite{Borji2013Quantitative}, mean absolute error (MAE) \cite{Willmott2005Advantages} and distribution-based metric similarity (SIM) \cite{Li2015A}. The comparative  DCT image signature (IS) and QDCT image signature (QIS) are computed to extensively validate the effectiveness of our aggregation signature (AS) method,
particularly on both MSRA-B \cite{Liu2007Learning} and small90 databases. There are  5000 images in MSRA-B, which is a large scale image database for quantitative evaluation of visual attention algorithms. From the results in Table \ref{2}, we observe that our method achieves overall better performance quantitatively than IS and QIS in terms of the MAE, NSS, SIM measures, and thus leading to a better estimation of the visual distance between the predicted saliency map and the ground truth. Fig. \ref{xianzhutu} provides the saliency maps of different methods, and the ground truth on images from small90, which shows that the background is more suppressed in aggregation signature with resepct to the others methods.
In terms of running speed, the aggregation signature module achieves 32 frames per second (FPS) in our experiments.

\begin{table}[]
\centering
\caption{Experiments on metrics (MAE, NSS, SIM) among IS, QIS and AS on MSRA-B and small90.
Bold fonts highlight the best performance.}
\label{my-label}
\begin{tabular}{c|lll|lll}
\hline
\multicolumn{1}{l|}{} & \multicolumn{3}{c|}{MSRA-B}                                                  & \multicolumn{3}{c}{small90}                                                 \\ \cline{2-7}
\multicolumn{1}{l|}{} & \multicolumn{1}{c}{MAE} & \multicolumn{1}{c}{NSS} & \multicolumn{1}{c|}{SIM} & \multicolumn{1}{c}{MAE} & \multicolumn{1}{c}{NSS} & \multicolumn{1}{c}{SIM} \\ \hline
IS                    & 0.2659                  & 0.9710                  & 0.3574                   & 0.1513                  & 3.6968                  & 0.0163                  \\
QIS                   & 0.2607                  & 0.9808                  & 0.3630                   & 0.1260                  & 4.4293                  & 0.0288                  \\
AS                    & \textbf{0.2559}         & \textbf{0.9844}         & \textbf{0.3695}          & \textbf{0.0660}         & \textbf{7.2658}         & \textbf{0.0611}         \\ \hline
\end{tabular}
\label{2}
\end{table}

\subsection{Aggregation Signature on Tracking}
 We empirically set the iteration number $R$ equal to 4, the saliency patches $M$ as 6. For other parameters, we follow the previous work \cite{Zhang2017Output} and set $\xi=1$, $ \tau=3$, $T_g=1.6$ in all experiments for fair comparisons.

\begin{figure}[!t]
\centering
\includegraphics[scale=0.42]{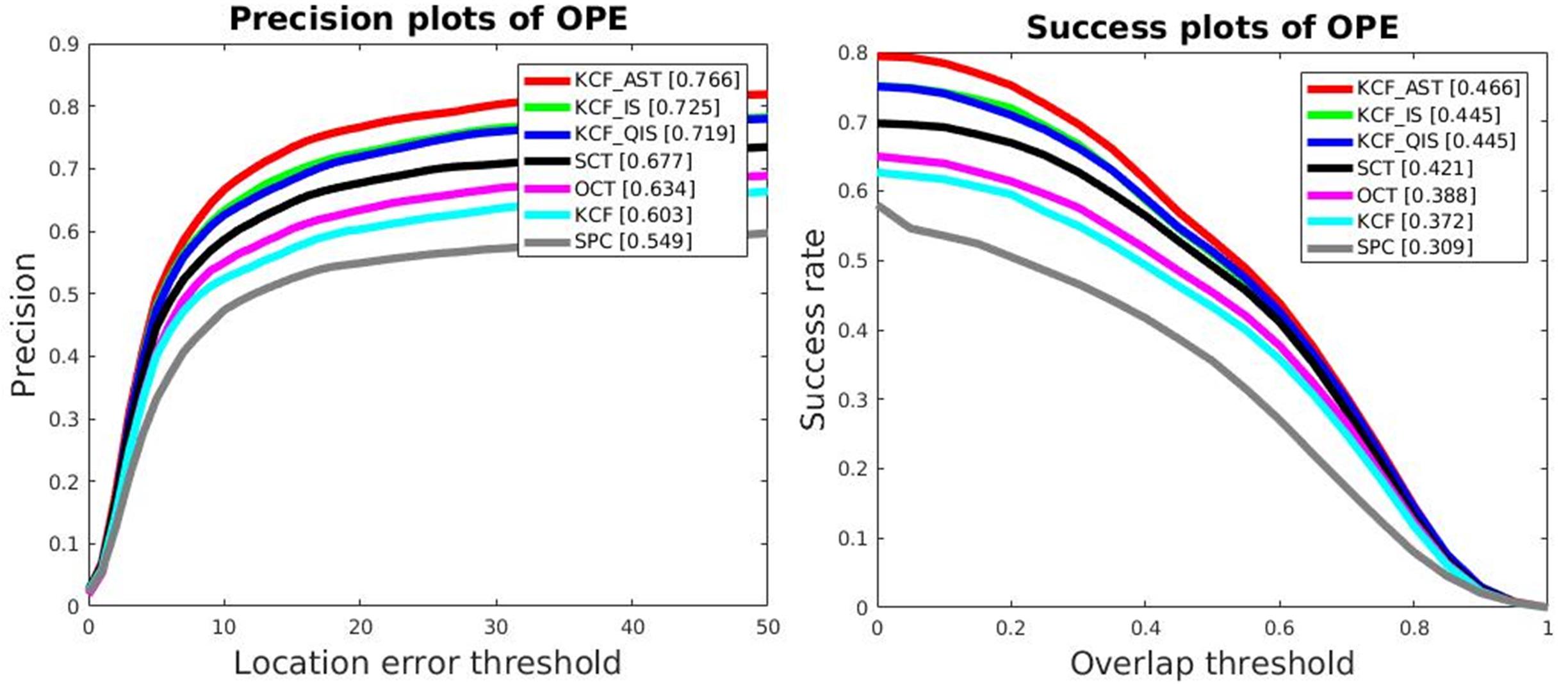}
\caption{Precision and success rate plots for AST performance on small90, in which SPC and SCT are the other two  saliency-based trackers, while KCF\_QDCT and KCF\_DCT are employed to compare the aggregation signature performance in tracking.}
\label{prsr4}
\end{figure}

We then test the performance of aggregation signature in tracking (AST) by comparing with DCT image signature, QDCT image signature, which are incorporated with KCF, based on small90. The results in Fig. \ref{prsr4} reveal that the aggregation signature clearly outperforms other signatures in small object tracking. Also, we use one-pass evaluation (OPE) \cite{Wu2015Object} to evaluate our results in the whole experiments section. Furthermore, we compare KCF\_AST with other saliency-based trackers, including saliency prior context model (SPC) \cite{Ma2017A} and structuralist cognitive tracker (SCT) \cite{Choi2016Visual} in the same figure. KCF\_AST (76.6\%) is about 22\% higher than SPC (54.9\%), and 9\% higher than SCT (67.7\%) in terms of the precision, while KCF\_AST (46.6\%) is about 16\% higher than SPC (30.9\%) and 5\% higher than SCT (42.1\%) based on the average success rate.

 We also compare our trackers with OCT, which also exploits the similar failure detection scheme to improve KCF. One can note that the performance of KCF\_AST is higher than OCT by 13.2\%  and 7.8\% in terms of  precision and  success rate, respectively.

\begin{figure}[!t]
	\centering
	\includegraphics[scale=0.24]{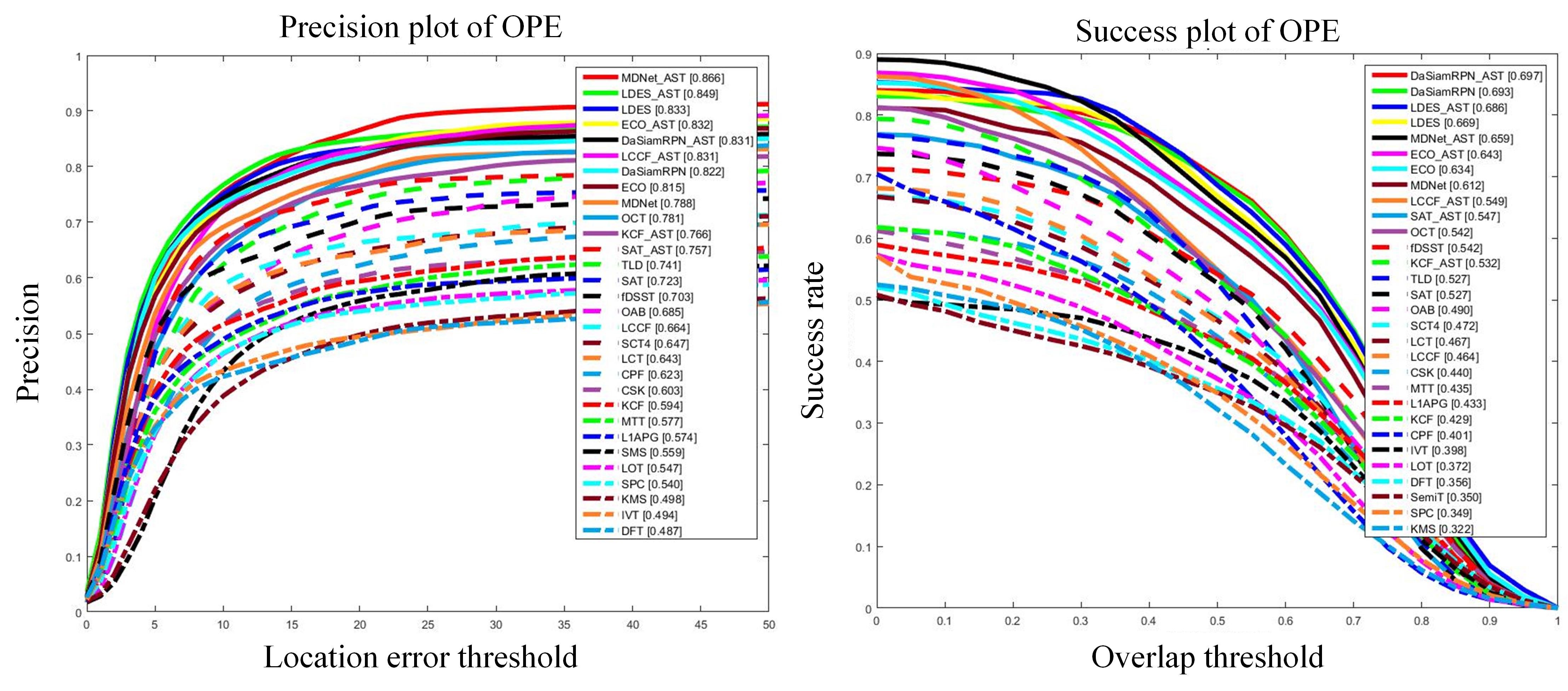}
	\caption{Precision and success rate plots on small90.}
	\label{prsr90}
\end{figure}

\textbf{The small90 benchmark: }In Fig. \ref{prsr90}, we further show the precision and success plots of 30 state-of-the-art trackers including SiamRPN \cite{Zhu_2018_ECCV} \cite{Li_2018_CVPR}, LDES \cite{Li2019Robust}, SAT \cite{T-IP}, TLD \cite{Kalal2012Tracking}, LCT \cite{Ma2015Long}, OCT \cite{Zhang2017Output}, CSK \cite{Rui2012Exploiting}, CT \cite{Zhang2012Real}, STC \cite{Zhang2014Fast}, KCF \cite{Henriques2015High}, ECO \cite{Danelljan2017ECO}, MDNet \cite{Nam2016Learning}, LCCF \cite{Zhang2017Latent}, SRDCF \cite{Danelljan2015Learning} and CPF \cite{Rez2002Color}, generated by the benchmark toolbox.
While several  baseline algorithms, e.g., LDES, DaSiamRPN, ECO, have shown promising potential in tracking small objects,  our AST still helps achieve the precision rates of 84.9\% (LDES\_AST), 83.1\% (DaSiamRPN\_AST), 83.2\% (ECO\_AST)  which improve its counterpart base trackers by 1.6\%, 0.9\%, 1.7\% respectively. Meanwhile, the above three trackers with our AST on achieves a success rate of 68.6\%, 69.7\%, 64.3\%, outperforming the base trackers by 1.7\%, 0.4\%, 0.9\% respectively.
Besides, our MDNet\_AST outperforms by 7.1\% and 4.0\% respectively to achieve a precision rate of 86.6\% and a success rate of 65.9\% compared to MDNet. This again confirms that our aggregation signature can consistently improve the performance of base trackers. Likewise, LCCF\_AST also shows a significant incremental performance, compared with the base tracker LCCF. Besides, when compared with the state-of-the-art re-detection trackers, our LCCF\_AST (54.8\%) significantly outperforms its base tracker LCCF (46.4\%), and also TLD (52.7\%), LCT(46.7\%) and OCT (54.2\%) by 2.1\%, 8.3\% and 0.7\% in terms of the success rate on small90, respectively.  The superior tracking performance confirms that our method is more effective than the state-of-the-art re-detection trackers such as TLD, LCT and OCT.

\begin{figure*}[!t]
	\centering
	\includegraphics[scale=0.92]{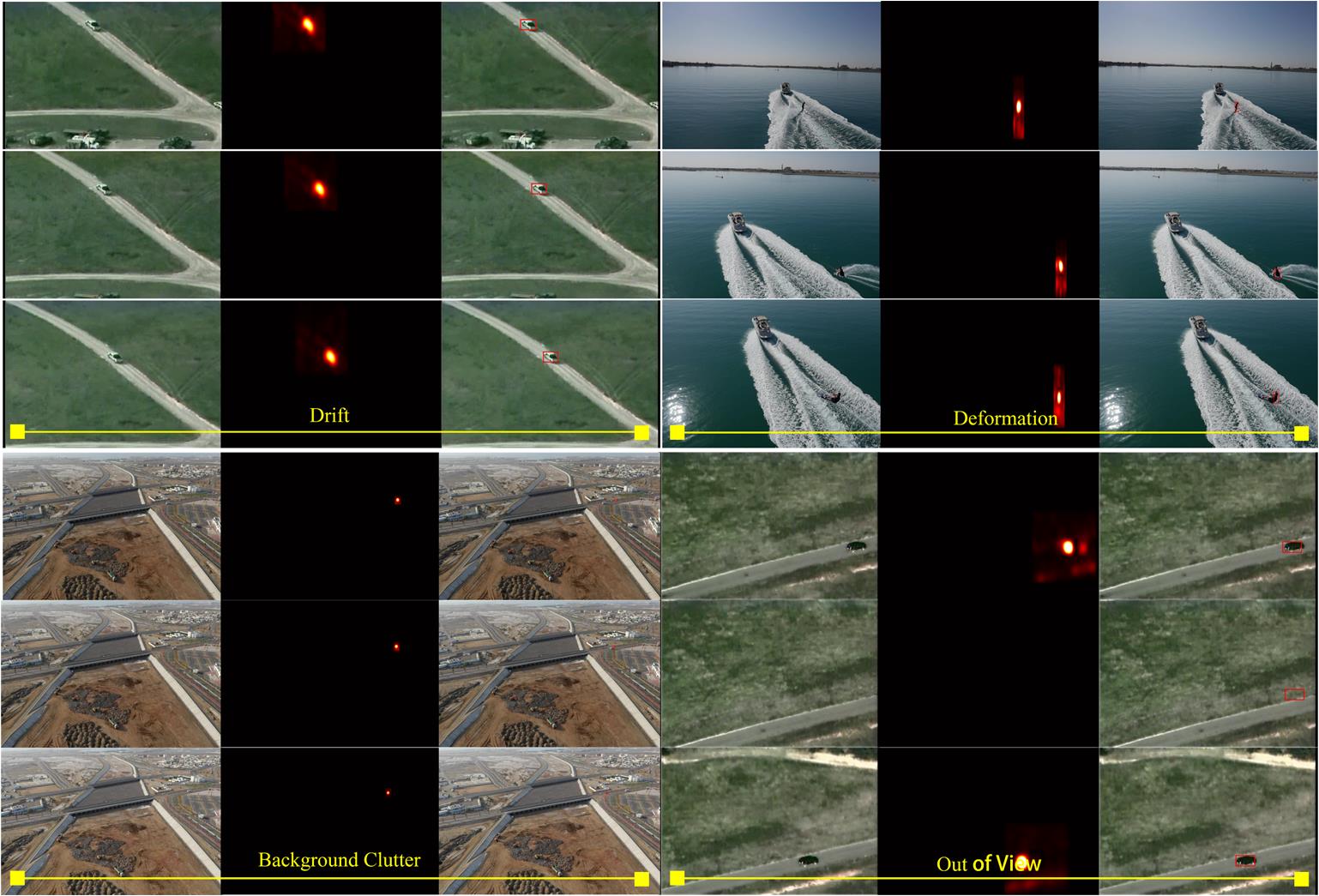}
	\caption{Representative tracking results on four challenging sequences (fastcar, wakeboard, truck and blackcar). For each subfigure, the current frame, the saliency maps obtained by Aggregation Signature (AS) and the corresponding tracking results are shown in the first, middle and right column, respectively. We can see that our AST tracker could tackle the drifting, deformation, background clutter and out-of-view challenge due to the usage of aggregation signature.}
	\label{keyframe}
\end{figure*}

We illustrate some examples for KCF\_AST in Fig. \ref{keyframe} to show how our aggregation signature helps to improve the tracking performance.
In the sequences selected from small90, the tracked objects are subject to severe image quality deterioration during the tracking process. In particular: 1) the background of the scene presents clutters while many objects are similar to the target in appearance and 2) severe drifting or long-time out of view results in directly drift of the target in the far range.  %
In addition, we adopt the MDNet, LCCF (deep feature) and the KCF as base trackers in our frameworks for comparision of visual tracking experiments. Results are shown in Fig.11; our main goal here is to show how our method helps to drastically reduce the tracking failure.

Observed from the results on Fig. \ref{keyframe} and Fig. 11, we can conclude that the aggregation signature can effectively improve the performance of base trackers, especially for small object tracking, and both saliency detection and tracking are enhanced by incorporating our image signature. As a final consideration, we acknowledge that the proposed method has the ability to relocate the target when drifting, and performs very well on the small target sequences.

\begin{figure}[!t]
\centering
\includegraphics[scale=0.23]{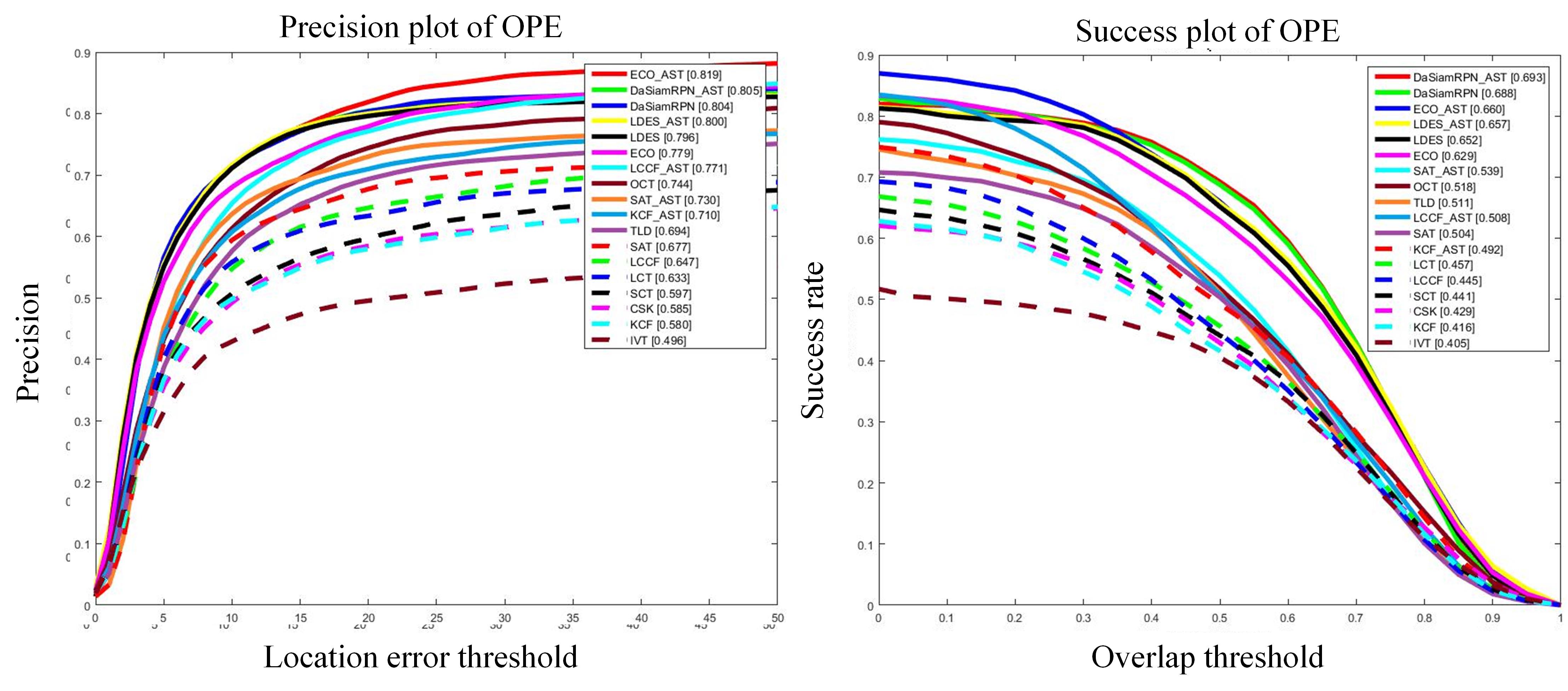}
\caption{Precision and success rate plots on small112.}
\label{small112}
\end{figure}

\begin{figure}[!t]
\centering
\includegraphics[scale=0.23]{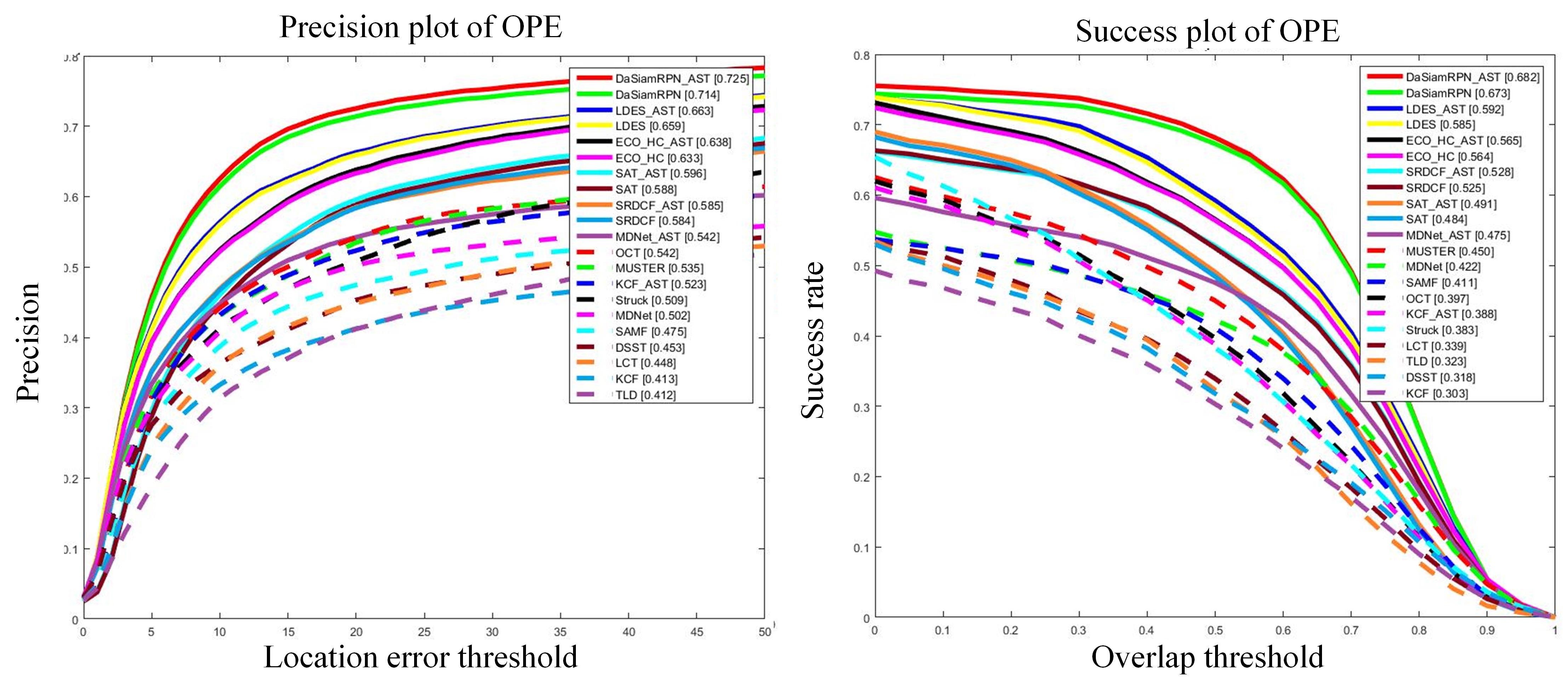}
\caption{Precision and success rate plots on UAV123\_10fps.}
\label{prsr123}
\end{figure}

{\textbf{The small112 benchmark: }We further collect a new benchmark dataset with 112 fully annotated sequences to facilitate the performance evaluation. On the basis of small90, the added 22 sequences are more difficult sequences.
As shown in Fig. \ref{small112},
KCF\_AST, LCCF\_AST, ECO\_AST improve the performance of KCF, LCCF, ECO from 58.0\%, 64.7\%, 77.9\% to 71.0\%, 77.1\%, 81.9\% on precision rate and 41.6\%, 44.5\%, 62.9\% to 49.2\%, 50.8\%, 66.0\% on success rate, which demonstrates that AST improves  these base trackers significantly on complex small object tracking sequences. Though the baseline  trackers, such as SiamRPN, LDES, perform very well, still 0.1\% and 0.4\% improvements on precision and 0.5\% and 0.5\% improvements on success rate have been obtained by AST, which validates the effectiveness of AST. Observed from the experimental results, all the trackers endowed with the aggregation signature module perform consistently better than the base trackers, which further validates the effectiveness of the proposed approach. Also, the results show that better base trackers gain less  performance improvements. The reason  might be that aggregation signature  is less useful if the drifting is not obvious, which is the case of using  a better tracker.

\textbf{The UAV123\_10fps benchmark: }We test ASTs on UAV123\_10fps \cite{Mueller2016A} as shown in Fig. \ref{prsr123}, which contains 123 sequences posing many challenges.
Compared to the base tracker MDNet, we can see the aggregation signature (MDNet\_AST) significantly improves  the performance of MDNet from 50.2\% to 54.2\% in precision rate and 42.2\% to 47.5\% in success rate, which further validates the effectiveness of the proposed method.  %
While KCF\_AST is about 6\% higher than KCF based on the precision, and is about 8\% higher based on success rate.
As for these more recent state-of-the-art trackers such as LDES, DaSiamRPN, ECO,  their corresponding ASTs  still achieve better results than these base trackers.

\begin{table*}[]
\centering
\caption{Precision and Success rate for the 11 attributes in Small90. Bold fonts highlight the best performance}
\label{my-label}
\begin{tabular}{c|c|c|c|c|c|c|c|c|c|c|c|c}
\hline
Precision    & SPC   & fDSST & OCT   & SCT   & KCF   & KCF\_AST & LCCF  & LCCF\_AST & MDNet & MDNet\_AST     & ECO   & ECO\_AST       \\ \hline
IV           & 0.396 & 0.619 & 0.715 & 0.551 & 0.491 & 0.707    & 0.538 & 0.765     & 0.707 & \textbf{0.798} & 0.719 & 0.747          \\
SV           & 0.495 & 0.710 & 0.723 & 0.618 & 0.574 & 0.805    & 0.706 & 0.805     & 0.775 & 0.794          & 0.768 & \textbf{0.809} \\
OCC          & 0.619 & 0.678 & 0.751 & 0.726 & 0.673 & 0.772    & 0.692 & 0.732     & 0.799 & \textbf{0.803} & 0.757 & 0.758          \\
DEF          & 0.542 & 0.706 & 0.767 & 0.676 & 0.599 & 0.757    & 0.671 & 0.805     & 0.807 & \textbf{0.844} & 0.777 & 0.793          \\
MB           & 0.303 & 0.491 & 0.631 & 0.421 & 0.353 & 0.582    & 0.390 & 0.684     & 0.516 & 0.717          & 0.696 & \textbf{0.726} \\
FM           & 0.353 & 0.573 & 0.746 & 0.500 & 0.412 & 0.645    & 0.452 & 0.789     & 0.573 & \textbf{0.809} & 0.770 & 0.803          \\
IPR          & 0.438 & 0.672 & 0.811 & 0.604 & 0.522 & 0.752    & 0.623 & 0.844     & 0.787 & \textbf{0.877} & 0.779 & 0.805          \\
OPR          & 0.464 & 0.704 & 0.838 & 0.625 & 0.551 & 0.782    & 0.650 & 0.869     & 0.831 & \textbf{0.921} & 0.808 & 0.833          \\
OC           & 0.237 & 0.374 & 0.880 & 0.327 & 0.293 & 0.611    & 0.431 & 0.795     & 0.721 & \textbf{0.855} & 0.494 & 0.664          \\
BC           & 0.533 & 0.696 & 0.770 & 0.655 & 0.599 & 0.769    & 0.653 & 0.815     & 0.786 & \textbf{0.855} & 0.789 & 0.804          \\
LR           & 0.578 & 0.717 & 0.816 & 0.666 & 0.625 & 0.783    & 0.697 & 0.858     & 0.805 & \textbf{0.900} & 0.845 & 0.863          \\ \hline
Success rate & SPC   & fDSST & OCT   & SCT   & KCF   & KCF\_AST & LCCF  & LCCF\_AST & MDNet & MDNet\_AST     & ECO   & ECO\_AST       \\ \hline
IV           & 0.209 & 0.379 & 0.423 & 0.328 & 0.291 & 0.422    & 0.322 & 0.445     & 0.430 & \textbf{0.487} & 0.451 & 0.464          \\
SV           & 0.264 & 0.459 & 0.396 & 0.361 & 0.324 & 0.416    & 0.393 & 0.439     & 0.511 & 0.519          & 0.504 & \textbf{0.524} \\
OCC          & 0.343 & 0.435 & 0.465 & 0.460 & 0.439 & 0.469    & 0.446 & 0.461     & 0.502 & \textbf{0.507} & 0.480 & 0.479          \\
DEF          & 0.305 & 0.454 & 0.456 & 0.425 & 0.378 & 0.460    & 0.411 & 0.477     & 0.524 & \textbf{0.540} & 0.508 & 0.514          \\
MB           & 0.150 & 0.299 & 0.396 & 0.260 & 0.201 & 0.368    & 0.230 & 0.402     & 0.324 & 0.464          & 0.453 & \textbf{0.474} \\
FM           & 0.185 & 0.367 & 0.473 & 0.317 & 0.246 & 0.421    & 0.282 & 0.481     & 0.374 & 0.537          & 0.514 & \textbf{0.538} \\
IPR          & 0.251 & 0.412 & 0.470 & 0.367 & 0.316 & 0.446    & 0.374 & 0.483     & 0.486 & \textbf{0.541} & 0.480 & 0.491          \\
OPR          & 0.262 & 0.433 & 0.481 & 0.376 & 0.329 & 0.460    & 0.386 & 0.495     & 0.514 & \textbf{0.570} & 0.500 & 0.510          \\
OC           & 0.150 & 0.263 & 0.408 & 0.209 & 0.181 & 0.382    & 0.242 & 0.436     & 0.425 & \textbf{0.512} & 0.329 & 0.427          \\
BC           & 0.305 & 0.451 & 0.471 & 0.416 & 0.376 & 0.476    & 0.407 & 0.493     & 0.511 & \textbf{0.552} & 0.526 & 0.532          \\
LR           & 0.334 & 0.469 & 0.499 & 0.414 & 0.382 & 0.475    & 0.417 & 0.507     & 0.527 & \textbf{0.587} & 0.561 & 0.571          \\ \hline
\end{tabular}
\label{3}
\end{table*}
\textbf{The UAV20L benchmark: }We also test ASTs on the well-known benchmark UAV20L \cite{Mueller2016A} as shown in Fig. \ref{prsr20}, where some of the tracked objects are very small. The state-of-the-art SRDCF is chosen as the base tracker, leading to our SRDCF\_AST. Apparently, SRDCF\_AST obtains better performances with respect to the state-of-the-art.
As compared to the base tracker SRDCF, we can see the aggregation signature (SRDCF\_AST) significantly improves  the performance of SRDCF from 50.7\% to 53.1\% in precision rate, which further validates the effectiveness of the proposed method.  %
LCCF\_AST is about 7\% higher than LCCF, while KCF\_AST is about 3\% higher than KCF based on the precision. %
In addition, LCCF\_AST and KCF\_AST, though showing no outstanding performance in terms of success rate, still achieved better results than their base trackers, respectively.
Furthermore, as for the more state-of-the-art trackers LDES and DaSiamRPN, we also show that LDES\_AST and DaSiamRPN\_AST improve their base trackers by a clear margin.
\begin{figure}[!t]
\centering
\includegraphics[scale=0.22]{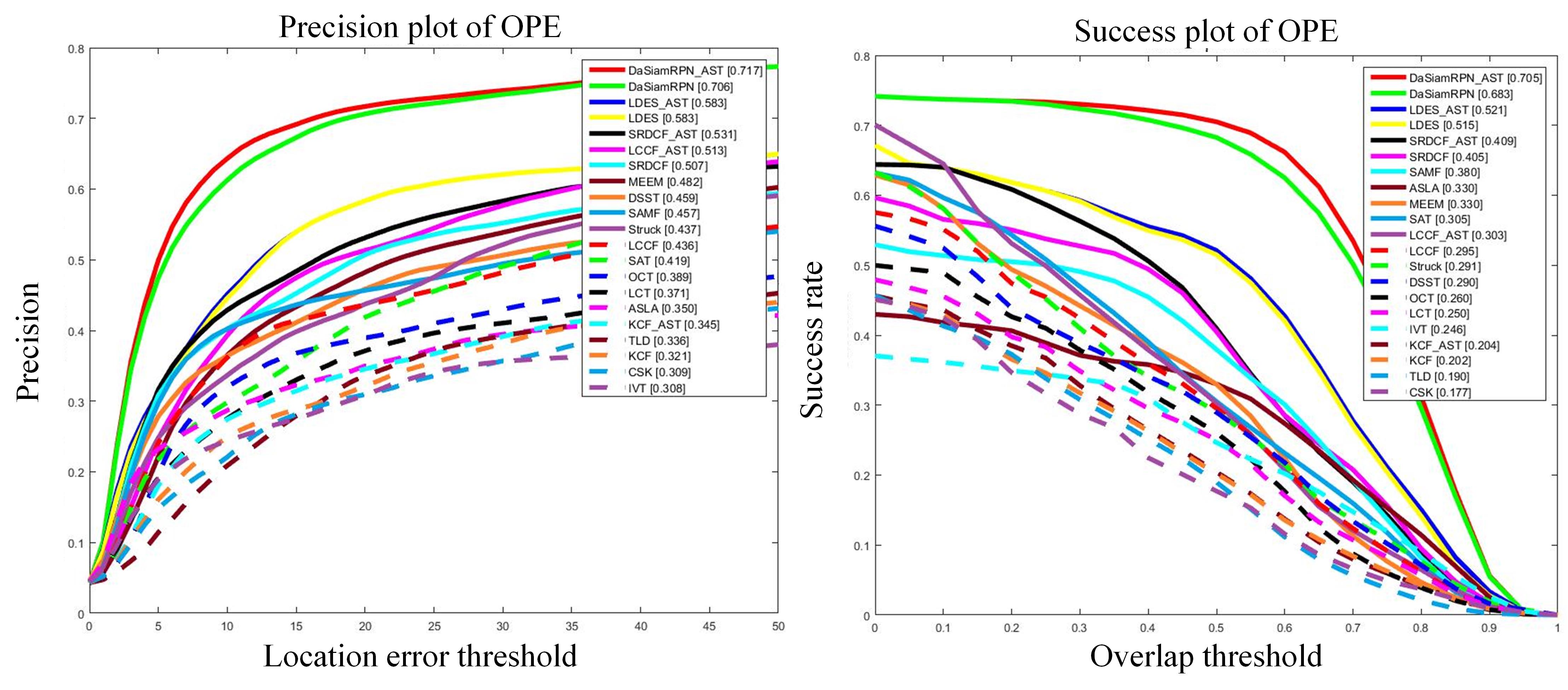}
\caption{Precision and success rate plots on UAV20L.}
\label{prsr20}
\end{figure}

\textbf{Quantitative Attribution Evaluation of Benchmarks: }
The full set of plots generated by the benchmark toolbox for small90 are also shown in Table \ref{3}. From the results, we can conclude that AST trackers achieve a much better performance in most cases for small-sized objects, especially for motion blur and fast motion, in which we can see all AST trackers improve dramatically, since saliency-based AST trackers can be more robust than base trackers to the variations mentioned previously.
To conclude,  AST  can consistently improve the results of base trackers in most cases, and AST-trackers achieve new state-of-the-art results.

\textbf{Speed analysis: }
In terms of tracking speed on small90, KCF\_AST has a processing rate of 120.88 frames per second (FPS), while LCCF\_AST based on deep features has 16.52 FPS, which show that our proposed trackers not only achieve the state-of-the-art results, but also performs in real time. Although the frame rate of the proposed tracking
framework has a drop, as compared to the original base tracker, the tracking performance is significantly improved on small90, e.g., 8.2\% improvement on LCCF in terms of success rate.

\section{Conclusions}
A new aggregation signature has been proposed to improve the small target tracking performance. The aggregation signature uses the target as a prior to adaptively locate the salient object, which is  deployed to re-detect the tracked objects when drifting. It is generic and can be used in conjunction with other trackers. We evaluated our tracking framework with KCF, SRDCF, LCCF, ECO, SAT, LDES, DaSiamRPN and MDNet. To validate the resulting aggregation signature tracker, we have also collected  new video datasets named small90 and small112, which contain fully annotated video sequences for small target tracking. The experimental results have clearly demonstrated how our methods improve the performance for the challenging situations, such as severe drifting, deformation and out of view. Furthermore, our approach will be extended to different applications in the future, such as large-scale retrieval \cite{Wu2018Unsupervised}\cite{Wu2019Joint} and classification \cite{Ding2019DECODE}.
\section{Acknowledgment}
The work was supported by the National Key Research and Development Program of China
(Grant No. 2016YFB0502602) and National Natural Science Foundation of China under Grant 61672079, in part by Supported by Shenzhen Science and Technology Program (No.KQTD2016112515134654).

\ifCLASSOPTIONcaptionsoff
  \newpage
\fi

\bibliographystyle{elsarticle-num}
\bibliography{reference}

\vspace{-11mm}
\begin{IEEEbiographynophoto}%
{Chunlei Liu} received her B.S. degree in Information Engineering from Nanjing University of Aeronautics and Astronautics, Nanjing, China, in 2016. She is now pursuing her phd degree at the Department of Electrical and information Engineering in Beihang University. Her research interests include computer vision and machine learning.
\end{IEEEbiographynophoto}
\vspace{-11mm}
\begin{IEEEbiographynophoto}%
{Wenrui Ding} received the doctorate degree in electrical and information engineering from Beihang University. She is currently in charge of information transmission and processing data link in the Unmanned System Research Institute in Beihang University. Her research interests include computer vision, the command and control of UAV, image processing, and pattern recognition.
\end{IEEEbiographynophoto}
\vspace{-11mm}
\begin{IEEEbiographynophoto}%
{Jinyu Yang} received her B.S. and M.Sc. degrees in Technology and Apparatus of Measuring and Control from Beihang University, Beijing, China and in Electronic Engineering from the Hong Kong University of Science and Technology, Hong Kong, in 2018 and 2019, respectively. She is now pursuing her Ph.D. degree in Computer Science in the University of Birmingham. Her research interests includes computer vision and machine learning.
\end{IEEEbiographynophoto}
\vspace{-11mm}
\begin{IEEEbiographynophoto}%
{Vittorio Murino}
 is full professor at the University of Verona, Italy, and director of PAVIS (Pattern Analysis and Computer Vision) department at the Istituto Italiano di Tecnologia. He took the Laurea degree in Electronic Engineering in 1989 and the Ph.D. in Electronic Engineering and Computer Science in 1993 at the University of Genova, Italy.
From 1995 to 1998, he was assistant professor at the Dept. of Mathematics and Computer Science of the University of Udine, Italy, and since 1998 he works at the University of Verona. He was chairman of the Department of Computer Science from 2001, year of foundation, to 2007, and coordinator of the Ph.D. program in Computer Science in the same university from 1999 to 2003. Prof. Murino is scientific responsible of several national and European projects, and evaluator of EU project proposals related to several frameworks and programs.
Since 2009, he is working at the Istituto Italiano di Tecnologia in Genova, Italy, leading the PAVIS department involved in computer vision, machine learning, pattern recognition and image analysis.
His main research interests include computer vision, pattern recognition and machine learning, more specifically, statistical and probabilistic techniques for image and video processing, with applications on (human) behavior analysis and related applications such as video surveillance, biomedical imaging, and bioinformatics.
Prof. Murino is co-author of more than 400 papers published in refereed journals and international conferences, member of the technical committees of important conferences (CVPR, ICCV, ECCV, ICPR, ICIP, etc.), and guest co-editor of special issues in relevant scientific journals.
He is also member of the editorial board of Computer Vision and Image Understanding, Machine Vision and Applications, and Pattern Analysis and Applications journals.
Finally, prof. Murino is IEEE Senior Member and IAPR Fellow.
\end{IEEEbiographynophoto}
\vspace{-11mm}
\begin{IEEEbiographynophoto}%
{Baochang Zhang} received the B.S., M.S. and Ph.D. degrees in Computer Science from Harbin Institue of the Technology, Harbin, China, in 1999, 2001, and 2006, respectively. From 2006 to 2008, he was a research fellow with the Chinese University of Hong Kong, Hong Kong, with Istituto Italiano di Tecnologia, Italy, and with Griffith University, Brisban, Australia. Currently, he is a tenured associate professor with Beihang University, Beijing, China. His current research interests include pattern recognition, machine learning, face recognition, and wavelets.
\end{IEEEbiographynophoto}
\vspace{-11mm}
\begin{IEEEbiographynophoto}%
{Jungong Han} is currently a tenured Data Science
Associate Professor with the University of Warwick, U.K. He has published more than 80 papers in the top venues including
IEEE/ACM Transactions and A* conference papers. His research interests
include computer vision, artificial intelligence, and machine learning.
\end{IEEEbiographynophoto}
\vspace{-11mm}
\begin{IEEEbiographynophoto}%
{Guodong Guo}  (M'07-SM'07) received the B.E. degree in automation from Tsinghua University,
Beijing, China, the Ph.D. degree in computer science from University of Wisconsin, Madison, WI, USA.
He is currently the Deputy Head of the Institute of Deep Learning, Baidu Research,
and also an Associate Professor with the Department of Computer Science and Electrical Engineering,
West Virginia University (WVU), USA. In the past, he visited and worked in several places, including INRIA,
Sophia Antipolis, France; Ritsumeikan
University, Kyoto, Japan; and Microsoft Research, Beijing, China;
He authored a book, {\em Face, Expression, and Iris Recognition Using Learning-based Approaches} (2008),
co-edited two books, {\em Support Vector Machines Applications} (2014) and {\em Mobile Biometrics} (2017),
and published over 100 technical papers. His research interests include computer vision, biometrics,
machine learning, and multimedia.
 He received the North Carolina State Award for
Excellence in Innovation in 2008, Outstanding Researcher (2017-2018, 2013-2014) at CEMR, WVU, and
New Researcher of the Year (2010-2011) at CEMR, WVU. He was selected the ``People's Hero
of the Week'' by BSJB under Minority Media and Telecommunications Council (MMTC) in 2013.
Two of his papers were selected as ``The Best of FG'13" and ``The Best of FG'15", respectively.
\end{IEEEbiographynophoto}

\end{document}